# Reduction of Overfitting in Diabetes Prediction Using Deep Learning Neural Network

Akm Ashiquzzaman[*], Abdul Kawsar Tushar[*], Md. Rashedul Islam[*, 1], and Jong-Myon Kim[**, 2]

[*] Department of CSE, University of Asia Pacific, Dhaka, Bangladesh
[**] Department of Electrical, Electronics, and Computer Engineering, University of Ulsan, Ulsan, Republic of Korea
{zamanashiq3, tushar.kawsar, rashed.cse}@gmail.com, jmkim07@ulsan.ac.kr

**Abstract.** Augmented accuracy in prediction of diabetes will open up new frontiers in health prognostics. Data overfitting is a performance-degrading issue in diabetes prognosis. In this study, a prediction system for the disease of diabetes is presented where the issue of overfitting is minimized by using the dropout method. Deep learning neural network is used where both fully connected layers are followed by dropout layers. The output performance of the proposed neural network is shown to have outperformed other state-of-art methods and it is recorded as by far the best performance for the Pima Indians Diabetes Data Set.

## 1 Introduction

Diabetes is a common physiological health problem among humans across gender, race and age. The term diabetic is applied when an individual is unable to break down glucose, for lack of insulin. The human organ called pancreas is responsible for generating the hormone called insulin, which is a very important enzyme that regulates the sugar level in human blood stream. It sanctions the human body to utilize sugar to generate energy; sans enough insulin, body cells cannot get the energy they need, consequently the sugar level in the blood gets too high, and many problems can emerge. Diabetes is not a curable disease; although, fortunately, it is treatable. Diabetes and related complications are responsible for the passing away of almost 200,000 Americans every year [1, 2]. In modern healthcare, predicting and properly treating diseases have become of foremost importance in medical prognostics fields. The whole process of determination of diabetes is completely manual, often suggested by the physician.

---

[1] Co-corresponding Author
[2] Corresponding Author

Smith et al. used the perceptron based algorithm called ADAPtive learning routine (ADAP), which is an early neural network model, to establish a diabetes prediction model for forecasting the arrival of diabetes mellitus. The system's performance measures were done using standard clinical benchmarks as specificity and sensitivity. The results obtained were then compared with those procured from applying linear perceptron models and logistic regression [3]. This method suffers from employing an early and complex structure of neural network which is responsible for performance degradation. On the other hand, Kayaer and Yıldırım proposed three separate neural network structures, which are multilayer perceptron (MLP), general regression neural network (GRNN), and radial basis function (RBF) and afterwards utilized the same data set to evaluate these three models. The performance gained by employing MLP was better than that of RBF method for all spread values tried. Among the three models evaluated, GRNN was able to provide the finest result on the test data [4]. Although not as complex as structures previously used, GRNNs are still convoluted structures; furthermore, this method, too, does not resort to any method for solving data overfitting.

The concept of deep learning is a fast-growing one which is teeming with ideas in recent years. Deep learning techniques are used in a range of diversified fields, including medical prognosis and optical character recognition [5]. In this paper, we will utilize the techniques of deep learning, namely - deep learning neural network, to propose a model for diabetes prognosis with high accuracy. This result is achieved with the help of a regularization layer called Dropout, which addresses the problem of overfitting arising from the use of deep fully connected layers.

The rest of this paper is structured as follows. Section 2 describes the proposed method, where the parts of the complete model are examined closely. Afterward, Section 3 delineates the dataset used in the experiment as well as the experiment procedures, and analyzes the results obtained via the proposed method. Then, Section 4 concludes the paper.

## 2 Proposed Method

Block diagram of the proposed method is outlined in Fig. 1. Here, the process is started by entering data into the input layer. Then there are two fully connected layers in place, each followed by a dropout layer. Finally, the decision is obtained from the output layer with a single node. Together these layers construct a multilayer perceptron, which is described in detail below.

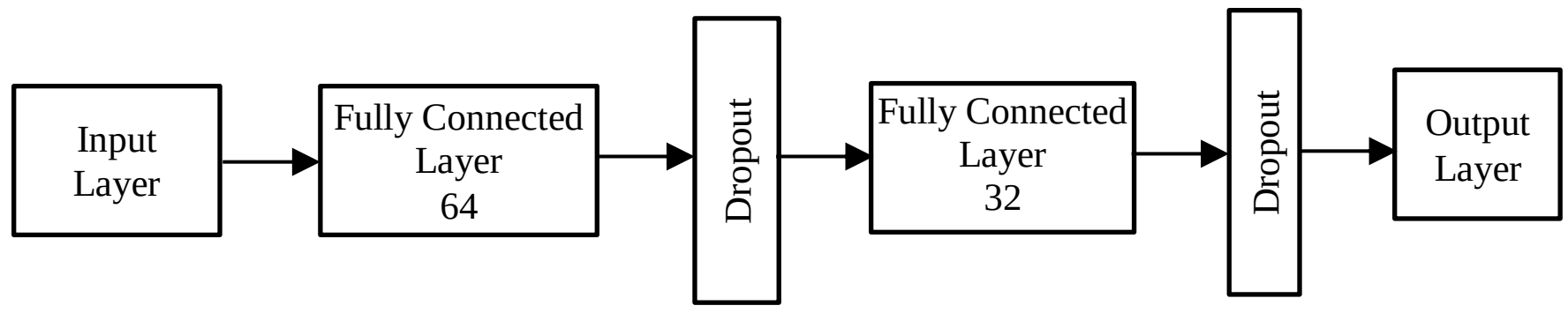

**Fig. 1.** Description of Proposed Method.

### 2.1 Multilayer Perceptron

Multilayer Perceptron is, in simple terms, a logistic regression classifier with hidden depth. Here, the input data is transformed with non-linearity, or activation functions to output one or more linearly separable classes. These intermediate layers are alluded to as hidden layers. A single hidden layer is sufficient to turn an MLP into a universal approximator. However, as it is learned, more hidden layers make the MLP more adaptive to the data [4].

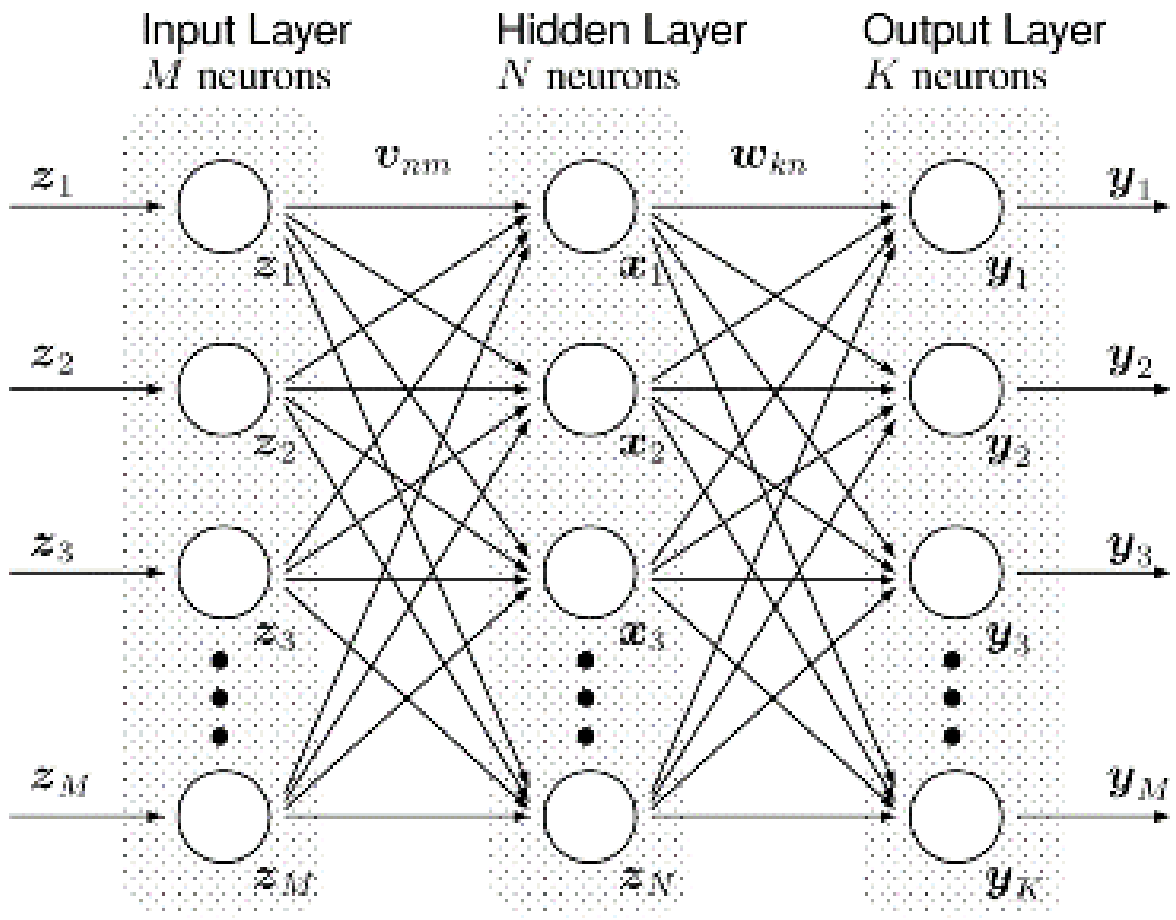

**Fig. 2.** A Feed Forward Neural Network/MLP.

Fig. 2 represents the common architecture of an MLP. In formal notation, an MLP with a sole hidden layer is a function $f\colon R^A \rightarrow R^B$, where $A$ and $B$ are respectively the sizes of input vector $x$ and output vector $f(x)$. Relation between input and output vectors can be coined as:

$$f(x) = \Phi\left(b^{(2)} + W^{(2)}\left(\varphi\left(b^{(1)} + W^{(1)}x\right)\right)\right) \quad \textbf{(1)}$$

with bias vectors $b^{(1)}$, $b^{(2)}$, weight matrices $W^{(1)}$, $W^{(2)}$, and activation functions $\Phi$ and $\varphi$. Here the activation function can be various mathematical threshold functions, i.e. *tanh(x)*, sigmoid, Exponential Linear Unit (ELU), or Rectified Linear Unit (ReLU). To train an MLP, we learn all parameters of the model, and to do that we use Stochastic Gradient Descent or any other relevant algorithm divided into mini-batches. The set of parameters to learn is the set $\theta = \{W^{(1)}, W^{(2)}, b^{(1)}, b^{(2)}\}$. Obtaining the gradients $\frac{\delta l}{\delta \theta}$ can be achieved through the backpropagation algorithm which is a special case of the chain-rule derivation [5].

### 2.2 Dropout

Dropout is nothing but a form of regularization. Srivastava et al. first implemented the dropout in their network to prevent overfitting problems of neural networks, which is depicted in Fig. 3 [6].

Dropout disables neurons in a neural network in such a random way that, during the learning phase, the network is forced to learn multiple representations of data. These representations are independent of each other and are derived from the same data. In this way, in various layers, neurons are hindered from co-adapting too well and this in turn reduces the possibility of overfitting. The DNN architecture uses a probability distribution in order to randomly exclude a number of neurons in each layer from updating weight. This results the neural network to learn from different representations. Krizhevsky et al. had implemented this regularization in their neural net in 2012, winning the prestigious Imagenet challenge of 2012 [7].

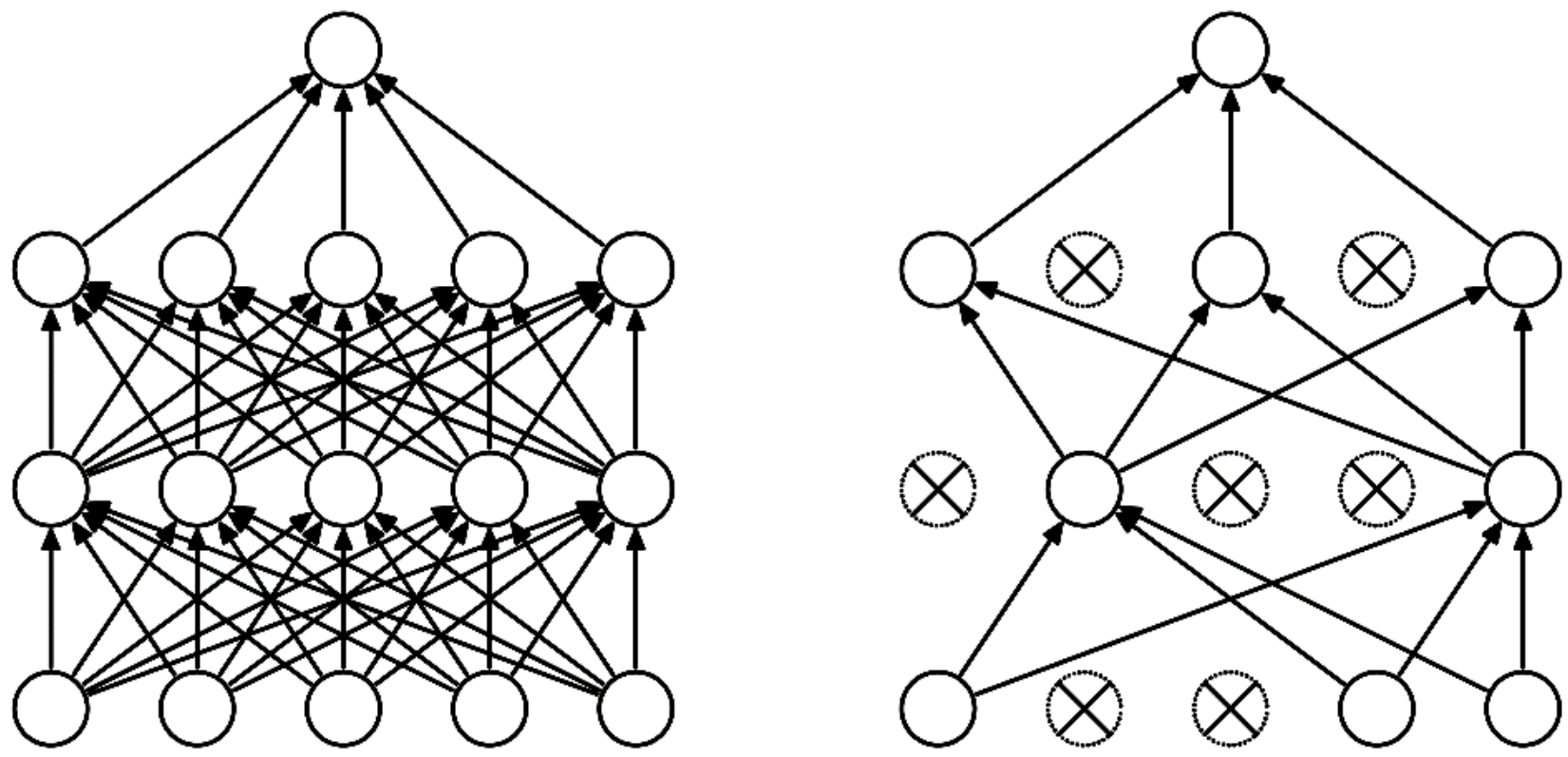

(a) Standard Neural Net (b) After Applying Dropout

**Fig. 3.** A demonstration of dropout, adapted from the method of Srivastava et al. [6].

The significant ability of the feed-forward DNN lies in its hidden layer. In our proposed method, we finalize on three layers altogether: the first input layer consists of 64 neurons, which is a multiple of the 8 input attributes. This layer utilizes ELU as its activation function. ELU is a special adaptive form of rectified linear unit which is chosen because of its sustainability toward vanishing gradient problem [8]. The next layer consists of 32 neurons, configured the same as the previous fully connected layer. The final layer is the one neuron output layer for the prediction which has Softplus function [9] as activation.

The numbers of nodes being accommodated in the hidden layers are decided by training several network configurations and choosing the optimal network topology from them. The criteria for optimality is fixed as yielding minimal seam square error (MSE) value and demonstrating the most increased predictability. Both the numbers of attributes as well the of output layer decision are in the powers of 2. Therefore, the numbers of nodes in both hidden layers are preferred to be powers of 2 as well. The point of note here is that, incorporating too many hidden layer neurons or hidden layers in the MLP architecture may sometimes lead to overfitting [10]. On the other hand, if the number of hidden layer neurons is relatively insufficient to capture the complexities of the problem, the issue of under fitting may arise [11].

Addition of dropout results in a generalized neural net rather than an overfitted one, in that it prevents the neurons from learning too much about the input data. This is done when dropout method forces some neurons in random order to be inactive during different phases of forward propagation as well as back-propagation. When neurons are randomly pitched out of the network during training, other neurons are forced to substitute for the missing neurons by handling the representation required to make predictions. This in turn results in multiple independent internal representations being learned by the network. Furthermore, this improves prediction results and validation scores [12, 13]. Each layer of the DNN is configured to have a dropout as a function in learning process. The first two layers of our proposed neural network has a low 25% probability in dropout, but the final layer has a 50% dropout rate to reduce overfitting.

Weights and biases of the entire network are first filled up according to a uniform function for maximum yield in learning. The Mean Squared Error (MSE) method is used as the loss function and Adadelta function is used for optimization. Back-propagation method is used for training the model via forward and backward passes. In forward passes, weights, biases, and inputs are combined to calculate a predicted value for each neuron. In backward passes, loss is calculated from difference of predicted and actual values, and that loss is used to update weights and biases in the model [5].

## 3 Experimental Results and Analysis

The Pima Indians Diabetes (PID) Data Set [14] is used in experiment. This data set is obtained from the UCI machine learning repository and is a subset of a bigger data set held by the National Institute of Diabetes and Digestive and Kidney Diseases [15]. The patients whose data are present in this database are women of Pima Indian inheritance who were older than 20 years of age and were residents of USA at the time of surveying. The binary output variable takes either 0 or 1, where 1 means testing positive and 0 is a testing negative for diabetes. 268 (34.9%) cases are present in class 1 for positive test and 500 (65.1%) cases in class 0 for negative test. There are eight clinical attributes and a brief overview is given in Table 1, along with specifications of the attributes in Table 2. As we can observe from the data in Table 2, the data could have been normalized. However, in deep learning, the DNN eventually learns the biases and filters the data accordingly. In our experiment, we use the whole dataset as it is, making no change to the data or any attributes.

The whole experiment is done in an Intel Core i5- 6200U CPU @ 2.30GHz 4 cores with 4 Gigabytes of DDR4 RAM. The MLP model is implemented with help of python Theano [16], with Keras [17] wrapper at the top.

**Table 1.** Attributes of PID Data Set [13].

| Class | Attribute Number |
|---|---|
| Pregnancy Count | 1 |
| Glucose concentration in plasma | 2 |
| Blood pressure (diastolic, mm Hg) | 3 |
| Thickness of triceps skin fold (mm) | 4 |
| 2-Hour serum insulin (µ U/ml) | 5 |
| Body mass index | 6 |
| Pedigree function of diabetes | 7 |
| Years of age | 8 |

**Table 2.** A Brief Description of PID Data Set [3].

| **Attribute Number** | **1** | **2** | **3** | **4** | **5** | **6** | **7** | **8** |
|---|---|---|---|---|---|---|---|---|
| **Mean** | 3.8 | 120.9 | 69.1 | 20.5 | 79.8 | 32.0 | 0.5 | 33.2 |
| **Standard Deviation** | 3.4 | 32.0 | 19.4 | 16.0 | 115.2 | 7.9 | 0.3 | 11.8 |
| **Minimum** | 0 | 0 | 0 | 0 | 0 | 0 | 0.078 | 2.42 |
| **Maxi-mum** | 17 | 199 | 122 | 99 | 846 | 67.1 | 2.42 | 81 |

The results of the previous methods discussed in the paper as well as result of the proposed method are depicted in Table 3. The proposed method has outperformed all the other methods previously described. Smith et al., with their proposed method, yields some good results, but the overall performance is not up to the mark. The specificity and sensitivity of their algorithm is 76% on the test set of 192 instances due to using methods that are now outdated. On the other hand, the three different methods proposed by Kayaer et al. have three superior yields during training phase; however, the accuracy drops significantly in the main cross-validation testing. This drop could be attributed to data overfitting. In contrast, the proposed method has the advantages of dropout as the regularization, which gives the network a considerable boost in performance. As a result, the overfitting issue that has been plaguing the other methods has minimal effect on the proposed method. Furthermore, the network is fed the data in raw format for processing, which is a distinct approach from previous methods. Due to this, the system can learn connections between the raw data values in a unique way. In the end, the proposed DNN had an 88.41% accuracy with 0.1 split validation, which is the new recorded accuracy for the PID dataset.

**Table 3.** Results of Different Methods Compared.

| ID no. | Method Name | Accuracy | Remarks |
|---|---|---|---|
| 1. | Smith et al. | 76% | Regression Network |
| 2. | Kayaer et al.<br>Training Set | 82.99% | DNN |
| 3. | Kayaer et al.<br>Test Set | 80.21% | Same DNN |
| 4. | Kayaer et al.<br>Mean Correct Prediction | 82.29% | GRNN |
| 5. | Proposed Method | 88.41% | DNN, with Dropout |

## 4 Conclusion

In this research, the application of dropout method is proposed in order to reduce data overfitting in predictive model. This model is used for forecasting the disease of diabetes. A novel form of deep neural network for diabetes prognosis with increased accuracy is discussed for this purpose. In this way, the accuracy achieved is 88.41% over the PID Data Set. By diminishing the effect of overfitting in the proposed model, increased accuracy is achieved via experimentation. As a result, performance for predictive models for diabetes can now have better prediction scores or performance gains which can lead to future breakthroughs in health prognostication.

ACKNOWLEDGMENT

This work was supported by the Korea Institute of Energy Technology Evaluation and Planning (KETEP) and the Ministry of Trade, Industry & Energy (MOTIE) of the Republic of Korea (No. 20162220100050 and No. 20161120100350). It was also funded in part by The Leading Human Resource Training Program of Regional Neo industry through the National Research Foundation of Korea (NRF) funded by the Ministry of Science, ICT and future Planning (NRF-2016H1D5A1910564), in part by the Basic Science Research Program through the National Research Foundation of Korea (NRF) funded by the Ministry of Education (2016R1D1A3B03931927), and in part by the development of a basic fusion technology in the electric power industry (Ministry of Trade, Industry & Energy, 201301010170D).

The authors acknowledge department of Computer Science and Engineering, University of Asia Pacific for supporting this research in various ways.